\documentclass{article}

\usepackage{amsmath}
\usepackage{amsfonts}
\usepackage{microtype}
\usepackage{graphicx}
\usepackage{subfigure}
\usepackage{booktabs} 

\usepackage{hyperref}

\usepackage[noend]{algorithmic}

\usepackage[accepted]{icml2018}

\icmltitlerunning{High Performance Zero-Memory Overhead Direct Convolutions}

\begin{document}

\twocolumn[
\icmltitle{ 
  High Performance Zero-Memory Overhead 
  Direct Convolutions 
}





\begin{icmlauthorlist}
\icmlauthor{Jiyuan Zhang}{cmu}
\icmlauthor{Franz Franchetti}{cmu}
\icmlauthor{Tze Meng Low}{cmu}

\end{icmlauthorlist}

\icmlaffiliation{cmu}{Department of Electrical and Computer Engineering, Carnegie Mellon University, Pittsburgh, USA}

\icmlcorrespondingauthor{Jiyuan Zhang}{jiyuanz@andrew.cmu.edu}

\icmlkeywords{High performance, convolution layer, deep learning, embedded systems, Internet of Things, zero memory, direct convolutions}

\vskip 0.3in
]



\printAffiliationsAndNotice{}  

\begin{abstract}
The computation of convolution layers in deep neural networks
typically rely on high performance routines that  trade
space for time by using additional memory (either for packing purposes
or required as part of the algorithm) to improve performance.  The
problems with such an approach are two-fold. First, these routines
incur additional memory overhead which reduces the overall size of the
network that can fit on embedded devices with limited memory capacity. Second, 
these high performance routines were not optimized for performing convolution, 
which means that the performance obtained is  usually less than conventionally expected.
In this paper, we demonstrate that direct convolution,
when implemented {\em correctly}, eliminates all memory overhead, and yields
performance that is between 10\% to 400\% times better than existing
high performance implementations of convolution layers on conventional and embedded CPU
architectures. We also show that a high performance direct convolution
exhibits better scaling performance, i.e. suffers less performance
drop, when increasing the number of threads.



\end{abstract}
\section{Introduction}
    \label{sec:intro}




\begin{figure}
    \centering
    \includegraphics[width=0.5\textwidth]{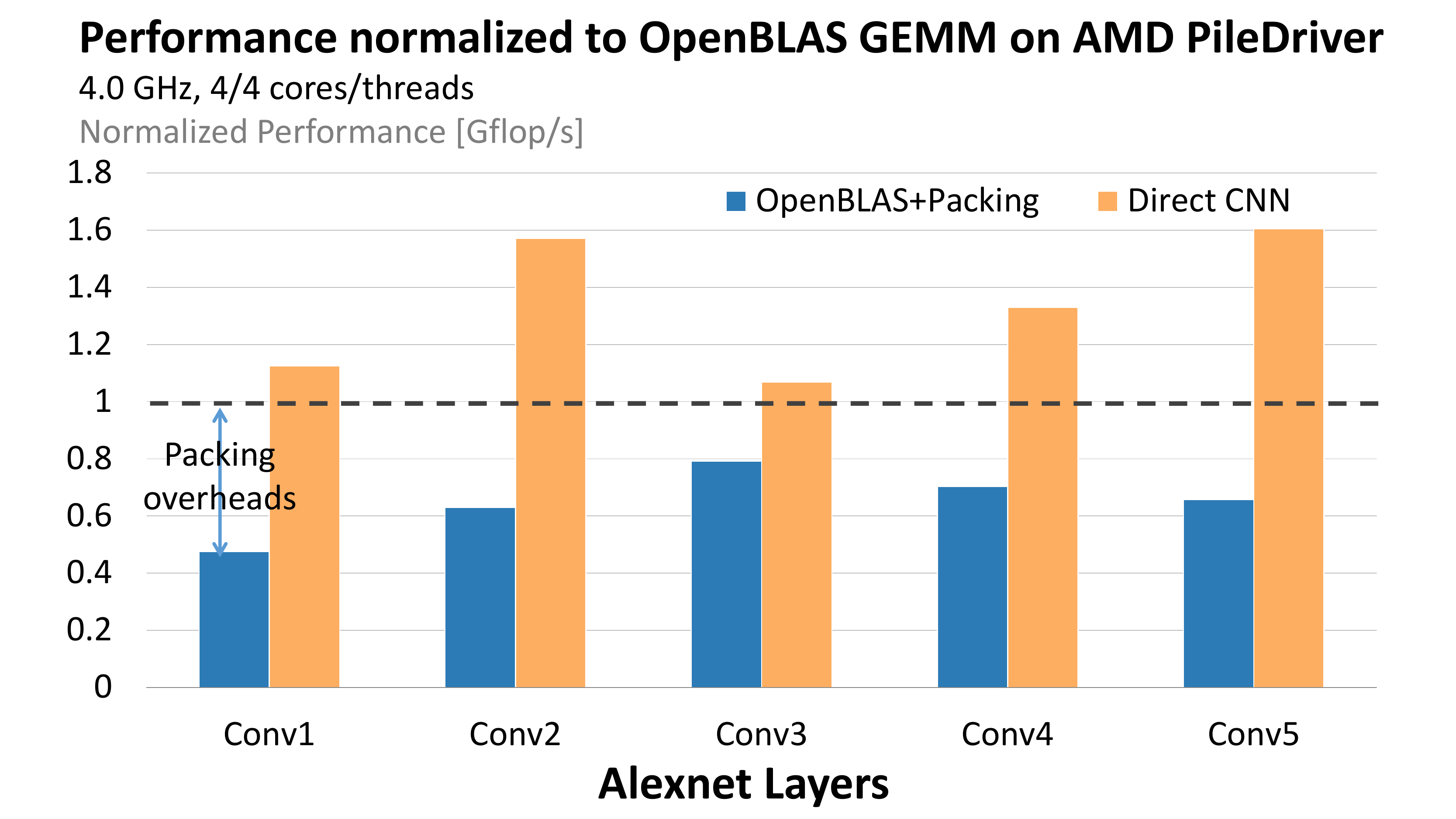}
    \vspace{-3mm}
    \caption{High performance direct convolution implementation
      achieves higher performance than a high performance matrix
      multiplication routine, whereas matrix-multiplication based
      convolution implementations suffers from packing overheads and
      is limited by the performance of the matrix multiplication
      routine}
\vspace{-5mm}      
    \label{fig:example}
\end{figure}

Conventional wisdom suggests that computing convolution layers found
in deep neural nets via direct convolution is not efficient. As such,
many existing methods for computing convolution
layers~\cite{jiacaffe,mec} in deep neural networks are based on highly
optimized routines (e.g. matrix-matrix multiplication) found in
computational libraries such as the Basic Linear Algebra Subprograms
(BLAS)~\cite{BLAS3}. In order to utilize the matrix-matrix multiplication
routine, these frameworks reshape and selectively duplicate parts of
the original input data (collectively known as packing); thereby incurring additional memory 
space for performance.

There are two problems with this approach:
First, the
additional work of reshaping and duplicating elements of the input
data is a bandwidth-bounded operation that incurs an additional, and non-trivial time
penalty on the overall system performance.
Second, and more importantly, matrices arising from convolution layers often have dimensions
that are dissimilar from matrices arising from traditional high
performance computing (HPC) application. As such, the matrix-matrix
multiplication routine typically does not achieve as good a performance on
convolution matrices as compared to HPC matrices. 

To illustrate these drawbacks of existing methods, consider the
4-thread performance attained on various convolution layers in AlexNet
using an AMD Piledriver architecture shown in
Figure~\ref{fig:example}. In this plot, we present performance of 1) a
traditional matrix-multiply based convolution implementation linked to
OpenBLAS{\footnote{OpenBLAS is an open-source implementation of the
    GotoBLAS algorithm, the de-facto algorithm for matrix
    multiplication on CPUs~\cite{GotoBLAS}.}}~\cite{OpenBLAS} (blue)
and 2) our proposed high performance direct convolution implementation
(yellow).  Performance of both implementations are normalized to the
performance of only the matrix-matrix multiplication routine (dashed
line). This dashed line is the performance attained by matrix-matrix
multiplication {\em if packing is free}.  Notice that the performance of
OpenBLAS + Packing achieves less than 80\% of the performance of
matrix multiplication itself. This implies that the packing routine
degrades the overall performance by more than 20\%. In contrast, our
custom direct convolution implementation yields performance that
exceeds the expert-implemented matrix-matrix multiplication routine, even if 
packing was free. In addition, we attained the performance 
{\em without} any additional memory overhead.

It is timely to revisit how convolution layers are computed as machine
learning tasks based on deep neural networks are increasingly being
placed on edge devices~\cite{speech, medical}. These devices are often
limited in terms of compute capability and memory
capacity~\cite{copro, embed}. This means that existing methods that
trade memory capacity for performance are no longer viable solutions
for these devices.  Improving performance and reducing memory overheads
also bring about better energy efficiency~\cite{zhang3d}. While many work have focused on reducing the
memory footprint of the convolution layer through the approximation~\cite{compression}, quantilization\cite{quantization}, or sparsification of the weights~\cite{han},
few work tackle the additional memory
requirements required in order to use high performance routines.

{\bf Contributions.} Herein lies the contributions of this paper:
\begin{itemize}
\vspace{-2mm}
\item {\em High performance direct convolution.} We show that a high
  performance implementation of direct
  convolution can out-perform a expert-implemented matrix-matrix
  multiplication based convolution in terms of amount of
  actual performance, parallelism, and reduced memory overhead. This demonstrates
  that that direct convolution is a viable means of computing convolution
  layers.  
  \vspace{-6mm}
\item {\em Data layouts for input/output feature maps and kernel weights.} 
We proposed new data layouts for storing the input, output and kernel weights
required for computing a convolution layer using our direct convolution algorithm. The space required for these new data layouts is identical to the existing data storage scheme for storing the input, output and kernel weights {\em prior} to any packing or duplication of elements.
\end{itemize}
\vspace{-2mm}

\section{Inefficiency of Non-direct Convolutions}
In this section, we highlight the inefficiency of computing
convolution with existing methods used in many deep learning
frameworks.

\subsection{Fast Fourier Transform-based Implementations}
Fast Fourier Transform (FFT)-based
implementations~\cite{vasilache2014fast,fftcnn} of convolution were
proposed as a means of reducing the number of floating point
operations that are performed when computing convolution in the
frequency domain. However, in order for the computation to proceed,
the kernel weights have to be padded to the size of the input image,
incurring significantly more memory than necessary, specially when the
kernels themselves are small (e.g. $3 \times 3$).  

Alternative
approaches have been proposed to subdivide the image into smaller
blocks or tiles~\cite{nnpack}. However, such approaches also require
additional padding of the kernel weights to a convenient size (usually
a power of two) in order to attain performance.  Even padding the kernel weights
to small multiples of the architecture register size (e.g. 8 or 16) will result 
in factors of 7 to 28 increase in memory requirement.
This additional
padding and transforming the kernel to the frequency domain can be
minimized by performing the FFT on-the-fly as part of the computation
of the convolution layer. This, however, incurs significant
performance overhead, especially on embedded devices, as we will show
in the performance section (Section~\ref{sec:experiment}).


\subsection{Matrix Multiplication-based Implementations}

\begin{figure*}
    \centering
    \includegraphics[width=0.9\textwidth]{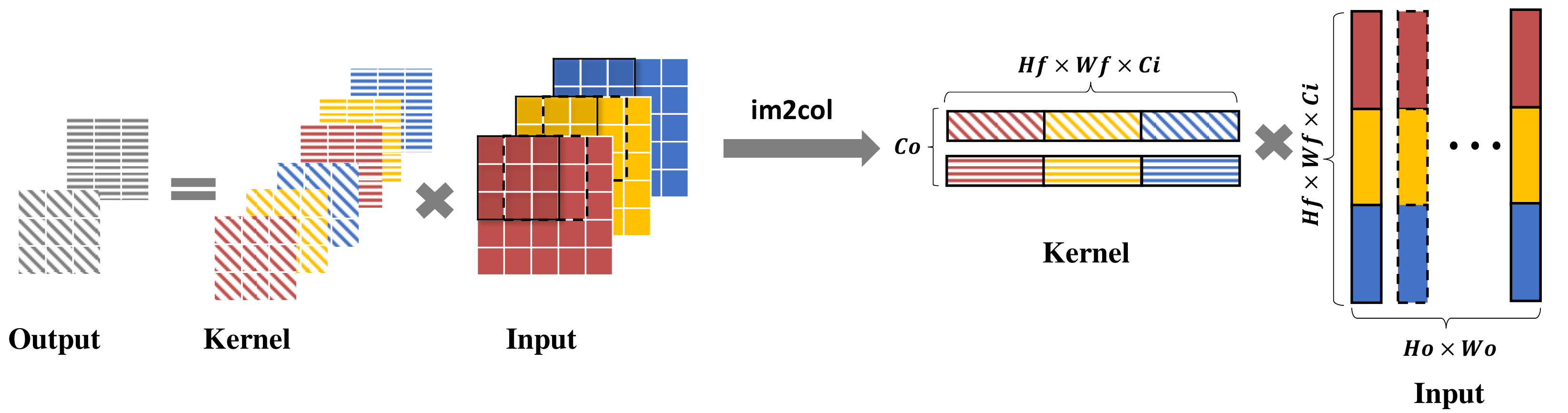}
    \vspace{-3mm}
    \caption{ The $5 \times 5 $ input image with 3 different channels
      (denoted with different colors) is convolved with two separate
      kernels to obtain a $3 \times 3 $ output with two output
      channels.  Packing is performed to turn three dimensional input
      images (left) into a two dimensional matrix (right) in order to
      utilize a high performance matrix multiplication routine. As
      $C_o$ and/or $ (H_o \times W_o)$ are often less than $H_f \times
      W_f \times C_i$, performance of standard matrix-matrix
      multiplication in many BLAS libraries are often sub-optimal. 
    }
    \label{fig:forward}
    \vspace{-3mm}
\end{figure*}

Another common approach is to cast the inputs (both the image and
kernel weights) into matrices and leverage the high performance
matrix-matrix multiplication routine found in the Level 3 Basic Linear
Algebra Subprogram (BLAS)~\cite{BLAS3} for computation. There are two
major inefficiencies with this approach:
\begin{itemize}
\vspace{-3mm}
\item {\em Additional memory requirements.} In order to cast the image
  into a matrix, a lowering operation is performed to cast the three
  dimensional image into a two dimensional matrix. Typically, this is
  performed via an operation conventionally called \texttt{im2col}
  that copies the $W_i \times H_i \times C_i$ image into a $(H_f
  \times W_f \times C_i) \times (H_o \times W_o)$ matrix which is then
  used as an input to the matrix-matrix multiplication call.  During
  this lowering process, appropriate elements are also duplicated. The
  additional memory required grows quadratically with the problem size~\cite{mec}.
  
Cho and Brand~\cite{mec} proposed an alternative lowering mechanism
that is more memory efficient by reducing the amount of duplication
required during the packing process. In their lowering routine, the
memory footprint is reduced by an average factor of 3.2 times over
\texttt{im2col}. This is achieved by eliminating the amount of
duplication required at the expense of additional matrix-matrix
multiplication calls.  Nonetheless, this is still an additional memory
requirement, and their computation still relies on a matrix-matrix
multiplication that is often sub-optimal for matrices arising from
convolution.

\item {\em Sub-optimal matrix matrix multiplication.}  In most BLAS
  libraries (e.g. GotoBLAS~\cite{GotoBLAS}, OpenBLAS~\cite{OpenBLAS},
  BLIS~\cite{BLIS1}), the matrix-matrix multiplication routine
  achieves the best performance when the inner dimensions, i.e. the dimension
  that is common between the two input matrices, of the input
  matrices are small compared to the overall dimensions of the output
  matrix. This particular set of matrix shapes is commonly found in
  scientific and engineering codes, for which these libraries are
  optimized. However, this particular set of shapes exercise only one
  out of six possible algorithms for matrix-matrix
  multiplication~\cite{GotoBLAS}.

  Recall that the \texttt{im2col} reshapes the input into a
$(H_f \times W_f \times C_i) \times (H_o \times W_o)$
matrix. This means that the inner dimensions of the input matrices are
often the larger of two dimensions (See Figure~\ref{fig:forward}). As
such, the performance of matrix matrix multiplication on this
particular set of input shapes is often significantly below the best
achievable performance. It has been shown that alternative algorithms
for computing matrix multiplications should be pursued for shapes
similar to that arising from convolution layers~\cite{ITXGEMM}.

Another reason that matrix-matrix multiplication is inefficient for
convolution layers is that parallelism in existing BLAS libraries are
obtained by partitioning the rows and columns of the input
matrices~\cite{BLIS3}. This partitioning of the matrices skews the
matrix shapes even farther away from the shapes expected by the
matrix-matrix multiplication routine. As such, the efficiency of the
routine suffers as the number of threads increases.
\vspace{-3mm}
\end{itemize}

\section{High Performance Direct Convolution}
\begin{algorithm}[tb]
   \caption{Naive Convolution Algorithm}
   \label{alg:naive}
\begin{algorithmic}
   \STATE {\bfseries Input:} Input $\mathcal I$, Kernel Weights $\mathcal F$, stride $s$; 
   \STATE {\bfseries Output:} Output $\mathcal O$
   \FOR{$i=1$ {\bfseries to} $C_i$}
   \FOR{$j=1$ {\bfseries to} $C_o$}
   \FOR{$k=1$ {\bfseries to} $W_o$}
   \FOR{$\ell=1$ {\bfseries to} $H_o$}
   \FOR{$m=1$ {\bfseries to} $W_f$}
   \FOR{$n=1$ {\bfseries to} $H_f$}
   \STATE {$ \mathcal O_{j,k,\ell} + \hspace{-1.2mm}= \mathcal I_{i, k\times s + m, \ell \times s + n} \times \mathcal F_{i,j,m,n} $}
   \ENDFOR
   \ENDFOR
   \ENDFOR
   \ENDFOR
   \ENDFOR
   \ENDFOR
\end{algorithmic}
\end{algorithm}

A naive implementation of direct convolution~(See
Algorithm~\ref{alg:naive}) is essentially six perfectly-nested loops
around a multiply-and-accumulate computational statement that computes
a single output element. Any permutation of the ordering of the loops
will yield the correct result. However, in order to obtain a high
performance implementation of direct convolution, it is essential that
these loops and their order are appropriately mapped to the given
architecture.  

\subsection{Strategy for mapping loops to architecture}
Our strategy for mapping the loops to a model architecture is similar
to the analytical model for high performance matrix-matrix
multiplication~\cite{BLIS4}. (1) We first introduce the model architecture 
used by high performance matrix-matrix
multiplication. (2) Next, we identify loops that utilize the
available computational units efficiently. (3) Finally, we identify the
order of the outer loops in order to improve data reuse, which in turn
will reduce the amount of performance-degrading stalls introduced into
the computation. In this discussion, we use the index variables show
in Algorithm~\ref{alg:naive} ($i, j, k, \ell, m, n$) to differentiate
between the loops.

\subsubsection{Model architecture}
We use the model architecture used the analytical model for high performance
matrix-multiplication~\cite{BLIS4}.  The model architecture is assumed
to have the following features:
\begin{itemize}
\vspace{-2mm}
 \item \textbf{Vector registers.}  We assume that our model
   architecture uses single instruction multiple data (SIMD)
   instruction sets. This means that each operation simultaneously
   performs its operation on $N_{\mbox{vec}}$ scalar output elements.  We
   also make the assumption that $N_{\mbox{vec}}$ is a power of two. When
   $N_{\mbox{vec}}$ is one, this implies that only scalar computations are
   available.  In addition, a total of $N_{\mbox{reg}}$ logical registers are
   addressable.
\vspace{-1mm}
\item \textbf{FMA instructions.}  We assume the presence of $N_{\mbox{fma}}$
  units that can compute fused multiply-add instructions (FMA). Each
  FMA instruction computes a multiplication and an addition. Each of
  these $N_{\mbox{fma}}$ units can compute one FMA instruction every cycle
  (i.e., the units can be fully pipelined), but each FMA instruction
  has a latency of $L_{\mbox{fma}}$ cycles. This means that $L_{\mbox{fma}}$ cycles
  must pass since the issuance of the FMA instruction before a
  subsequent dependent FMA instruction can be issued.
\vspace{-1mm}
\item \textbf{Load/Store architecture.} We assume that the architecture
is a load/store architecture where data has to be loaded into registers
before operations can be performed on the loaded data. On architectures
with instructions that compute directly from memory, we assume that those
instructions are not used.
\end{itemize}

\subsubsection{Loops to saturate computations}
The maximum performance on our model architecture is attained when all
$N_{\mbox{fma}}$ units are computing one FMA per cycle. However, because each
FMA instruction has a latency of $L_{\mbox{fma}}$ cycles, this means that
there must at least be $L_{\mbox{fma}}$ independent FMA instructions issued
to each computational unit. As each FMA instruction can compute
$N_{\mbox{vec}}$ output elements, this means that
\begin{equation}
\mathcal E \geq N_{\mbox{vec}}N_{\mbox{fma}}L_{\mbox{fma}},
\end{equation}
where $\mathcal E$ is the minimum number of independent output
elements that has to be computed in each cycle in order to reach the
maximum attainable performance.

Having determine that at least $\mathcal E$ output elements must be
computed in each cycle, the next step is to determine the arrangement
of these output elements within the overall output of the convolution
layer.  Notice that the output has three dimensions ($H_o \times W_o
\times C_o$) where $H_o$ and $W_o$ are primarily a function of the
input sizes, while $C_o$ is a design parameter of the convolution
layer.  Since $\mathcal E$ must be a multiple of $N_{vec}$, i.e. a
power-of-two, and $C_o$ can be chosen (and is the case in practice) to
be a power-of-two, {\em the $j$ loop is chosen as the inner-most
  loop}.

As the minimum number $\mathcal E$ is highly dependent on the number
and capability of the FMA computation units, we want to ensure that
there are sufficient output elements to completely saturate
computation. As such, {\em the $k$ loop that iterates over the
  elements in the same row of the output image is chosen to be the
  loop around the $j$ loop} {\footnote{It should be noted that the
    choice of $W_o$ over $H_o$ is arbitrary as the analysis is
    identical.}}.


\subsubsection{Loops to optimize data reuse}
The subsequent loops are ordered to bring data to the computational
units as efficiently as possible.

Recall that the inner two loops ($j$ and $k$) iterate over multiple
output elements to ensure that sufficient independent FMA operations
can be performed to avoid stalls in the computation units. As our
model architecture is a load/store architecture, this means that these
output elements are already in registers. Therefore, we want to bring
in data that allows us to accumulate into these output elements.

Recall that to compute a single output element, all $H_f \times W_f
\times C_i$ weights are multiplied with the appropriate element from
the input image and accumulated into the output element. This
naturally means that {\em the next three loops in sequence from the
  inner-most to outer-most are the $i, m, n$ loops.} This order of the
loops is determined based on the observation that the input of most
convolution layers is the output of another convolution layer. This
means that it would be advisable if data from both the input and
output are accessed in the same order. As such, we want to access the
input elements in the channels ($i$) before rows ($n$), which gives
us the $i, n, m$ ordering of the loops.

Having decided on five of the original six loops, this means that {\em
  outermost loop has to be the $l$ loop}. This loop traverses over the
remaining through different rows of the output. The original loop
order as shown in Algorithm~\ref{alg:naive} ($i, j, k, l, m, n$) is
transformed to the ($l, n, m, i, k, j$) loop ordering as shown in Algorithm~\ref{alg:ordered}.

\begin{algorithm}[tb]
   \caption{Reorder Convolution Algorithm}
   \label{alg:ordered}
\begin{algorithmic}
   \STATE {\bfseries Input:} Input $\mathcal I$, Kernel Weights $\mathcal F$, stride $s$; 
   \STATE {\bfseries Output:} Output $\mathcal O$
   \FOR{$\ell=1$ {\bfseries to} $H_o$}
   
   \FOR{$n=1$ {\bfseries to} $H_f$}
   \FOR{$m=1$ {\bfseries to} $W_f$}
   \FOR{$i=1$ {\bfseries to} $C_i$}
   \FOR{$k=1$ {\bfseries to} $W_o$}
   \FOR{$j=1$ {\bfseries to} $C_o$}
   
   \STATE {$ \mathcal O_{j,k,\ell} + \hspace{-1.2mm}= \mathcal I_{i, k\times s + m, \ell \times s + n} \times \mathcal F_{i,j,m,n} $}
   \ENDFOR
   \ENDFOR
   \ENDFOR
   \ENDFOR
   \ENDFOR
   \ENDFOR
\end{algorithmic}
\end{algorithm}

\vspace{-2mm}
\subsubsection{Blocking for the memory hierarchy}
{\bf Register Blocking.}The astute reader will recognize that we have
conveniently ignored the fact that $\mathcal E$, the number of minimum
output elements required to sustain peak performance,is upper bounded
by the number of registers as described by the following inequality:
\begin{equation}
\mathcal E  \leq N_{\mbox{reg}}{N_{\mbox{vec}}}.
\end{equation}
This upper bound imposed by the number of available registers means
that at most $N_{\mbox{reg}}N_{\mbox{vec}}$ elements can be kept in the
registers. This means that instead of iterating over all $C_o \times
W_o$ elements, loop
blocking/tiling~\cite{Wolfe:tiling} with block sizes of $C_{o,b}${\footnote{
$C_{o,b}$ is chosen to be a multiple of the vector length $N_{vec}$ so that
SIMD instructions can be better used for computation.}} and
$W_{o,b}$ has to be applied to the two inner-most loops to avoid
register-spilling that will degrade performance.


Applying loop blocking to the original $j$ and $k$ loops decomposes a
row from each of the output channel into smaller output images, each
of which having a row width and output channel of $W_{o,b}$, and
$C_{o,b}$ respectively. Since loop blocking decomposes the overall convolution
into smaller convolutions, the loop ordering previously described remains
applicable. However, we now need to determine how to traverse over the smaller
convolutions.

The loops $j'$ and $k'$ iterate over the blocks in the channel and row 
dimensions of the output, respectively. In addition, loops $jj$ and $kk$ iterate
with the respective blocks of channels and rows. We make the observation accessing input elements in the same row will require us to also access kernel weights in the same row. This suggest that the ordering of the loop should be similar to the loops traversing across the kernel weights. As such, the $k'$ loop is nested between
$\ell$ and $n$ loops. The $j'$ loop is set to be the outermost loop since it
is a parallel loop that facilitates parallelization.

{\bf Cache Blocking.} On architecture with more levels in the memory
hierarchy, i.e. architectures with caches, we can further partition
the input dataset into smaller partitions such that they fit into the
appropriate levels of the cache. Recall that the loops around $jj$ and
$kk$ accumulates $H_f \times W_f \times C_i$ intermediate results into
the output stored in the register.  Since $H_f$ and $W_f$, i.e. the
size of the kernel weights, are typically smaller than $C_i$, we
choose to partition the $i$ loop which iterates over $C_i$ input
channels for the next level in the memory hierarchy. 

The final algorithm for high performance direct convolution is shown
in Algorithm~\ref{alg:final}.

\begin{algorithm}
   \caption{Parallelized Direct Convolution Algorithm} 
   \label{alg:final}
\begin{algorithmic}
   \STATE {\bfseries Input:} Input $\mathcal I$, Kernel Weights $\mathcal F$, stride $s$; 
   \STATE {\bfseries Output:} Output $\mathcal O$
   \FOR{$j' =1$  {\bfseries to} $C_o / C_{o,b}$ {\bfseries in Parallel}}
   \FOR{$i'=1$   {\bfseries to} $C_i / C_{i,b}$ }
   
   \FOR{$\ell=1$   {\bfseries to} $H_o$ }
   
   \FOR{$k'=1$   {\bfseries to} $W_o / W_{o,b}$}
   
   \FOR{$n=1$   {\bfseries to} $H_f$}
   \FOR{$m=1$   {\bfseries to} $W_f$}
   \FOR{$ii=1$  {\bfseries to} $C_{i,b}$}
   
   \FOR{$kk=1$  {\bfseries to} $W_{o,b}$}
   \FOR{$jj =1$ {\bfseries to} $C_{o,b}$}
   \STATE {
   $
   \begin{array}{l}
   \mathcal O_{j'C_{o,b}+jj,{k'W_{o,b}}+kk,\ell}\quad+\hspace{-1mm}= \\
   \quad\mathcal I_{{i'C_{i,b}}+ii, {sk'W_{o,b}}+kk + m, \ell s + n}~\times \\
   \quad\mathcal F_{{i'C_{i,b}}+ii, j'\times C_{o,b}+jj,m,n}
  \end{array} 
  $}
   \ENDFOR
   \ENDFOR
   \ENDFOR
   \ENDFOR
   \ENDFOR
   \ENDFOR
   \ENDFOR
   \ENDFOR
   \ENDFOR
\end{algorithmic}
\end{algorithm}
\vspace{-3mm} 

\subsection{Parallelism}
In order to identify possible parallel algorithms, we first make the
observation that all output elements can be computed in
parallel. Since the output is a three dimensional object ($H_o \times
W_o \times C_o$), this means that parallelism can be extracted in at
least three different dimensions. 

Our direct convolution implementation extracts parallelism in the
output channel ($C_o$) dimension. Each thread is assigned a block of
output elements to compute, where each block of output elements is of
size $H_o \times W_o \times C_{o}/p$, where $p$ is the number of
threads used.

\section{Convolution-Friendly Data Layout}
We proposed new data layouts for the input and kernel data so that
data is accessed in unit stride as much as possible. This improves
data access and avoids costly stalls when accessing data from lower
levels of the memory hierarchy.  A key criteria in revising the layout
is that the output and the input image should have the same data
layout. This is because the input of most convolution layers is the
output of another convolution layer. Keeping them in the same data
layout will avoid costly data reshape between convolution
layers. However, to ensure compatibility with original input images,
we do not impose the proposed layout on the inputs to the first
convolution layer.


\subsection{Input/Output Layout}
We want to access the output data in unit stride. Therefore, we
determine the output data layout by considering how the elements are
accessed using the loop ordering shown in Algorithm~\ref{alg:final}.
Data accessed in the inner loops should be arranged closer together in
memory than data accessed in the outer loops. 

Five loops ($j,k, \ell, kk, jj$) iterate over the output data, which
suggests a five-dimensional data layout. However, this
is sub-optimal if we were to use it for the input data. This is because
$W_f$ elements in an input row is required
to compute one output element.  With the five-dimensional layout, 
a row of the input is blocked into blocks of $W_{o,b}$ elements. This 
 means that output elements that require input elements from two 
separate $W_{o,b}$  blocks will incur a large penalty as these
input elements are separated over a large distance in memory. As such 
we do not layout the data according to the $kk$ loop.

The proposed input/output layout is shown in
Figure~\ref{fig:layout}~(left).  The output data is organized into
sequential blocks of $H_o \times W_o \times C_{o,b}$, where in each
block, elements are first laid out in the channel dimension, before
being organized into a $H_o \times W_o$ row-major-order matrix of
pencils of length $C_{o,b}$.

\subsection{Kernel Layout}
Similar to the input/output layout, we use the loop ordering to
determine how to order the kernel weights into sequential
memory. Notice that the $\ell, k', kk$ loops in Algorithm~\ref{alg:final}
iterates over the height and width of the output in a single output
channel. As all output elements in the same output channel share the
same kernel weights, these loops provide no information as to how the
kernel weights should be stored. As such, we only consider the remaining
six loops.

\begin{figure*}
\centering
\begin{tabular}{cc}
\centering{\includegraphics[scale=0.35]{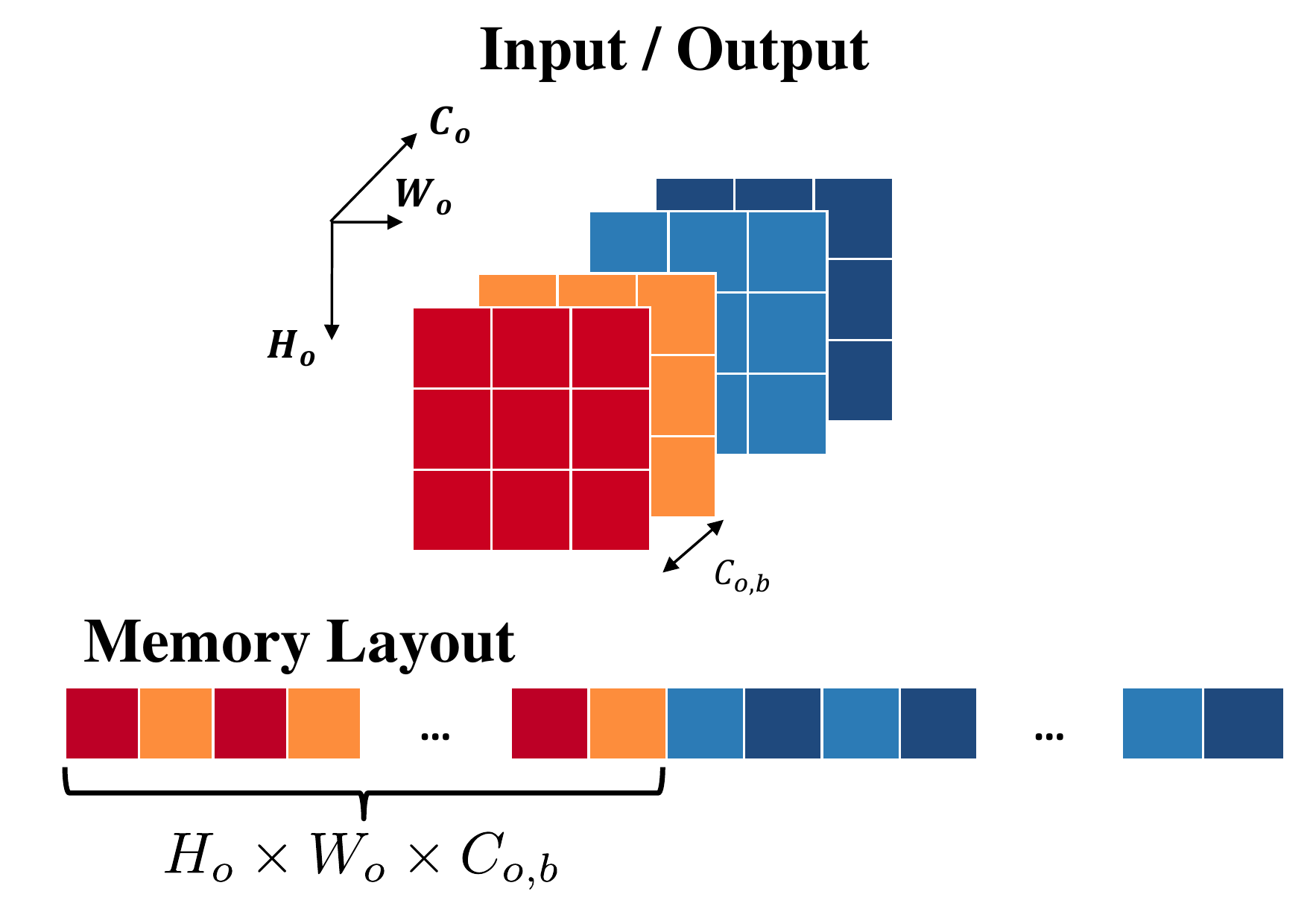}} &
\centering{\includegraphics[scale=0.25]{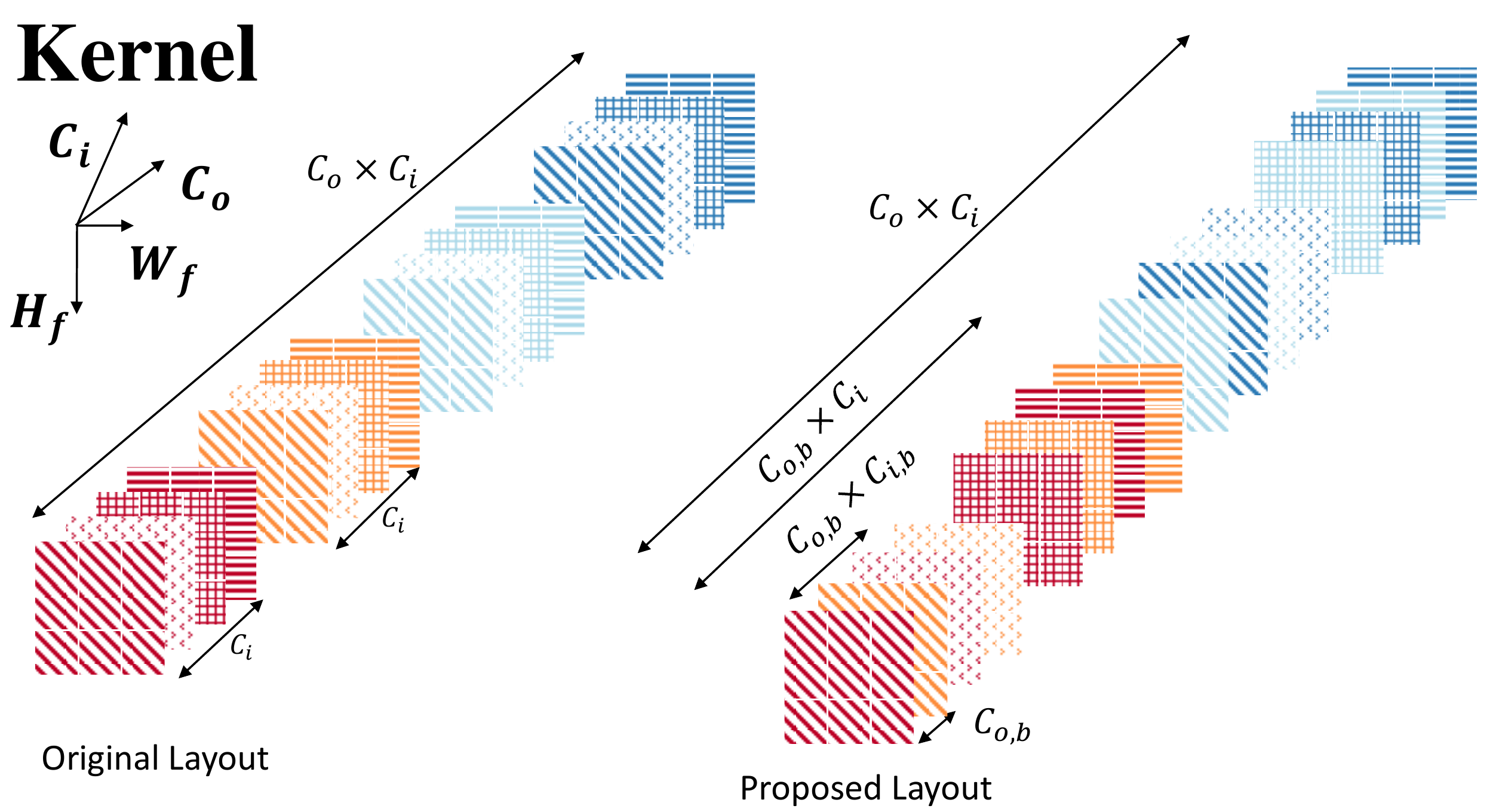}} 
\end{tabular}
\vspace{-3mm}
\caption{Convolution-friendly layout for input/output (left) and
  kernel weights (right). The output data is organized into sequential
  blocks of $H_o \times W_o \times C_{o,b}$, where in each block,
  the fastest dimension is in the channel dimension, followed by the column and 
  row dimension of the output. The kernel weights are organized into blocks of $H_o \times W_o \times C_{o,b} \times C_{i,b}$. The fastest dimension is the blocked output channel, followed by the blocked input channels, kernel width and height, input channels and then the output channels. }
\vspace{-5mm}
\label{fig:layout}
\end{figure*}

The kernel layout proposed by the remaining six loops is shown in
Figure~\ref{fig:layout}~(right). The fastest dimension in the kernel
layout is the blocked output channel ($C_{o,b}$) dimension, and is
dictated by the inner-most loop. The remaining dimensions from fastest to 
slowest are the blocked input channel ($C_{i,b}$), followed by the columns ($W_f$) and rows ($H_f$) of the kernel, the input channels ($C_{i}/C_{i,b}$) and finally 
the output channels ($C_o / C_{o,b}$).


\subsection{Backward compatibility} 
Given the successful deployment of convolution neural nets (CNN)in the
field, the proposed change in data layout will mean that trained
networks are unable to directly benefit from our proposed direct
convolution implementation. However, in order for a trained network to
use our proposed algorith, there is only a one-time cost of
rearranging the kernel weights into the proposed data layout.
Other network layers such as skip layers~\cite{Resnet}, and activation
layers are point-wise operations that should not require any
significant change in the implementation. Nonetheless, reordering the
loops used to compute these layers will likely yield better
performance.

\section{Results}
    \label{sec:experiment}



%

In this section, we present performance results of our direct CNN implementation against existing convolution approaches  on a variety of architecture. A mix of traditional CPU architectures (Intel and AMD) and embedded processor (ARM) found on embedded devices are chosen.

\vspace{-3mm}
\subsection{Experimental Setup}
{\bf Platform}
We run our experiments on Intel Core i7-4770K, AMD FX(tm)-8350, ARM Cortex-A57 architectures. The architecture details of those platforms are shown in Table .  

\begin{table}
\caption{Details of specific architectures used}
 \label{table:dims}
\begin{center}
\begin{small}
\begin{tabular}{ lccc }
 \toprule
  &Intel  & AMD & ARM  \\
  &i7-4770K & FX(tm)-8350 & Cortex-A57\\
 \midrule
 Architecture  & Haswell & Piledriver & ARMv8 \\
 Frequency & 3.5GHz & 4GHz & 1.1GHz\\
 Cores     & 4      & 4    &  2 \\
 $N_{vec}$ & 8      & 8  & 4 \\ 
 \bottomrule
\end{tabular}
\end{small}
\end{center}
\end{table}

\textbf{Software.}  We implement our direct convolution using
techniques from the HPC community~\cite{handsoff}.  We compare
performance our direct convolution implementation against
matrix-multiplication based convolution linked to high performance
BLAS libraries.  For matrix-multiplication based convolution, the
input data is first packed into the appropriate matrix using Caffe's
\texttt{im2col} routine before a high performance single-precision
matrix-multiplication (\texttt{SGEMM}) routine is called. The
\texttt{SGEMM} routine used is dependent on the architecture. On Intel
architecture, we linked to Intel's Math Kernel Library
(MKL)~\cite{MKL}, while OpenBLAS~\cite{OpenBLAS} is used on the other
two architectures. 
We also provide comparison against the FFT-based convolution
implementation provided by NNPACK~\cite{nnpack}, a software library
that underlies the FFT-based convolutions in Caffe 2~\cite{caffe2}. As
NNPACK provides multiple FFT-based (inclusive of Winograd)
implementations, we only report performance attained by the best
(fastest) implementation. We use the benchmark program supplied by 
NNPACK to perform our tests.


\textbf{Benchmarks.}
All implementations were ran against all convolution layers found in AlexNet~\cite{alexnet}, GoogLeNet~\cite{googlenet} and VGG~\cite{vgg}. The different convolution layers in these three CNNs span a wide range of sizes of input, output and kernel weights. They are also commonly used as benchmarks for demonstrating the performance of convolution implementations.

%
%
%

\subsection{Performance}
\begin{figure*}
\centering{\includegraphics[width=0.9\textwidth]{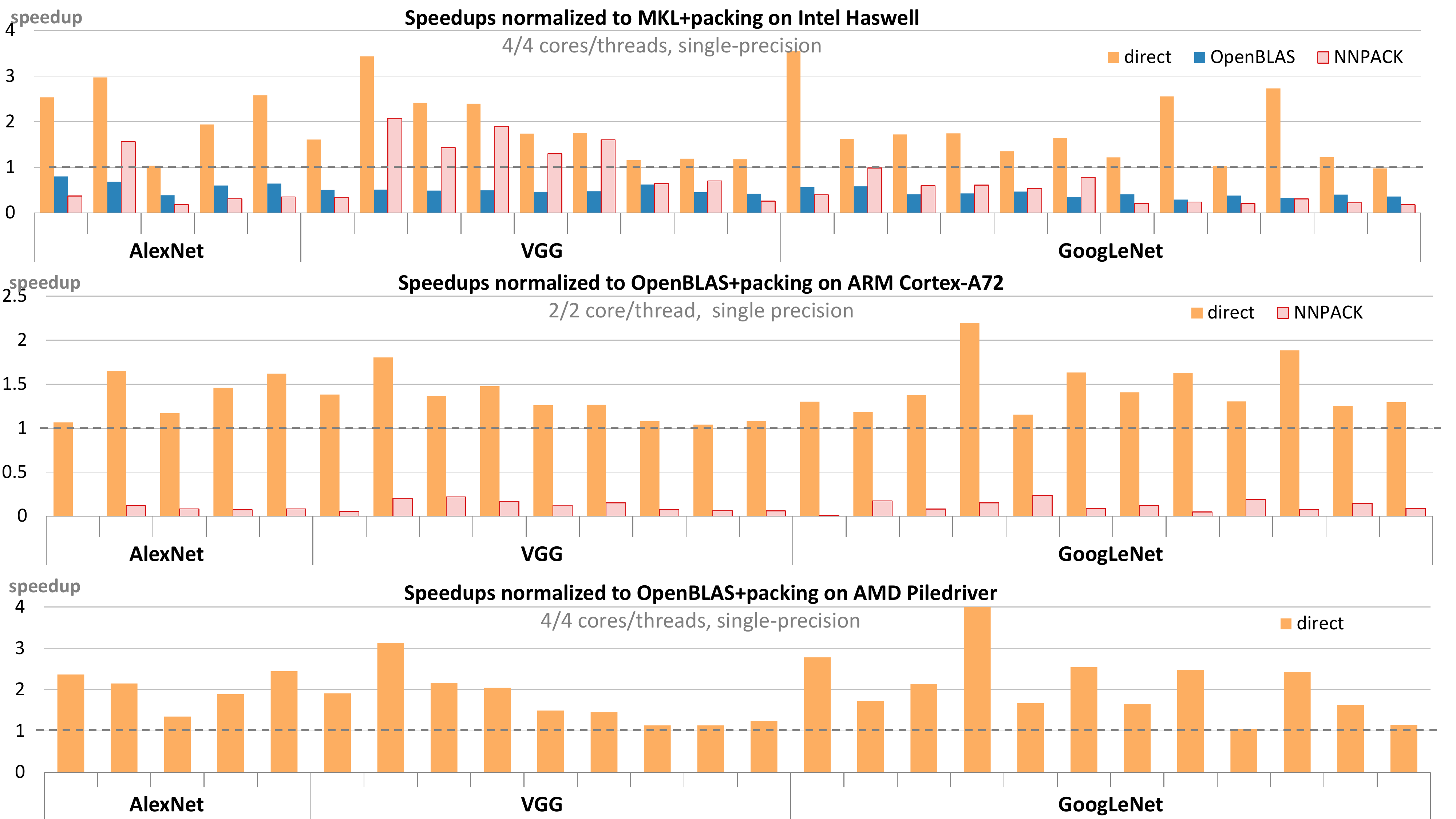}}
\caption{Performance of direct convolution 
  against existing high performance FFT-based and
  \texttt{SGEMM}-based convolution implementations. Performances
  of all implementations are normalized to the performance of 
  \texttt{SGEMM} +\texttt{im2col} routine. Direct convolution is highly 
  competitive against all other implementations achieving between 10\% and 400\% improvement in performance even against a BLAS library (Intel MKL) that optimizes for matrix shapes arising from convolution layers.}
\vspace{-3mm}
\label{fig:perf}
\end{figure*}

The relative performance of the different implementations normalized
to the \texttt{SGEMM}+ packing method are shown in
Figure~\ref{fig:perf}. Our direct convolution implementations 
out-performs all \texttt{SGEMM}-based convolutions on all
architectures by at least 10\% and up to 400\%. Our direct convolution out-performs 
\texttt{SGEMM} even when the BLAS library (MKL) optimizes for the appropriate matrix shapes arising from convolution. Against a BLAS library (OpenBLAS) that only optimizes for HPC matrices, we see a minimum of 1.5 times performance gain on 4 threads.

In comparison with the FFT-based implementations provided by NNPack,
the direct convolution implementation significantly out-performs  FFT-based
implementations for all layers on the ARM. 
As FFTs are known
to be memory-bandwidth bound, we suspect that the FFT may be the
bottleneck on a smaller architecture such as the ARM where available
bandwidth may be limited. On the Intel architecture, the results are
similar with direct convolution outperforming FFT-based
implementations. However, in this case the FFT-based implementations are able to out-peform
the \texttt{SGEMM}-based approach only when the dataset is
``sufficiently large'' to amortize the cost of performing the FFT itself. The AMD architecture is
not supported by NNPACK.

\begin{figure*}
\centering{\includegraphics[width=0.9\textwidth]{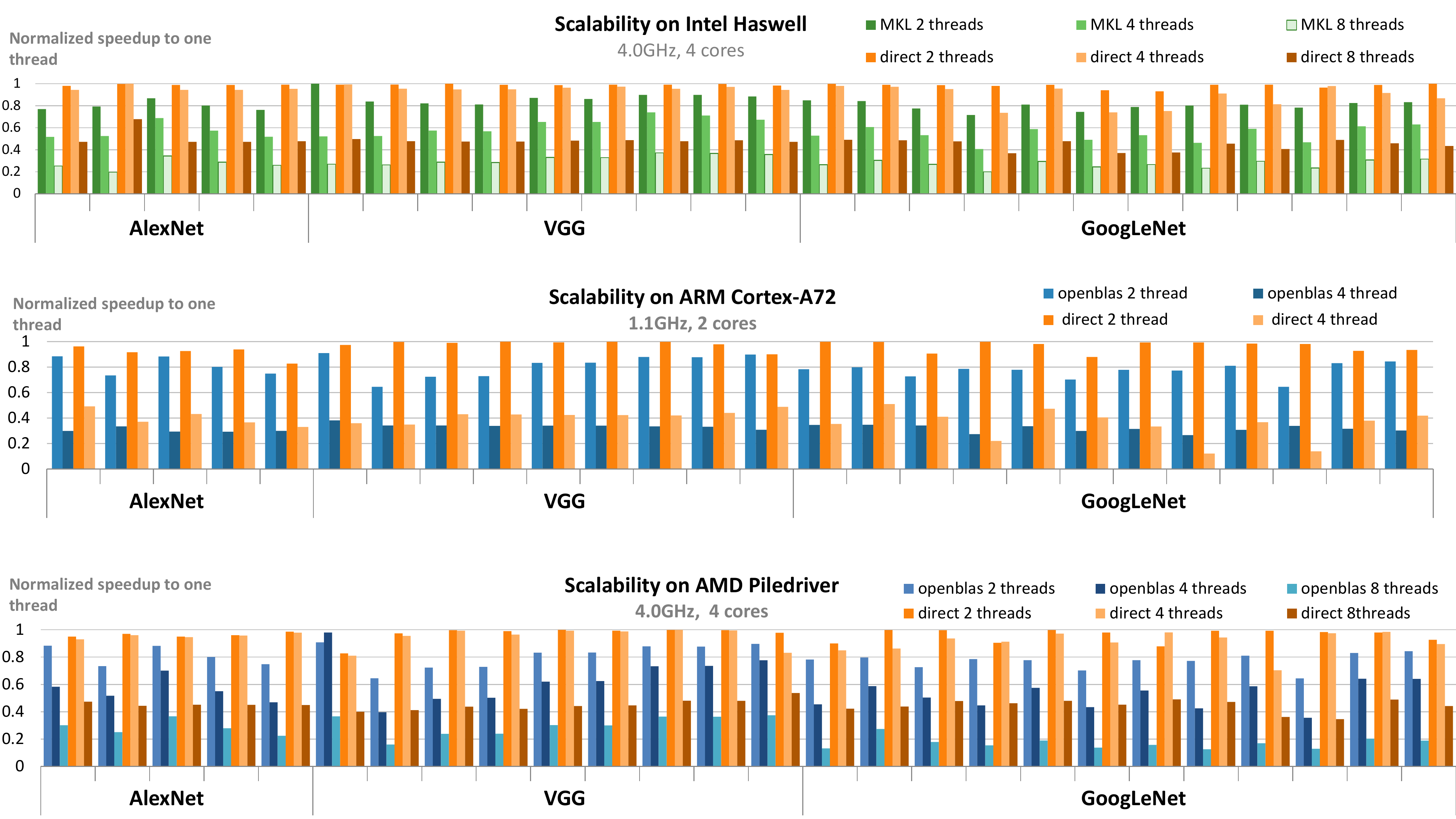}}
\caption{Scaling behavior with increasing number of threads. Our
  direct convolution implementation retains high GFLOPs per core
  performance as we increase the number of threads from 1 to the
  number of available cores. This is indicative of an efficient
  parallelized algorithm. When the number of threads exceeds the
  number of cores, excessive contention results in a significant drop
  in performance per core. In contrast, \texttt{SGEMM} has poor
  scalability even when the number of threads is low (e.g. 2).}
\vspace{-3mm}
\label{fig:parallel}
\end{figure*}

\vspace{-2mm}
\subsection{Parallel Performance}
In Figure~\ref{fig:parallel}, we compare the scalability of our
convolution performance by parallelizing the implementation with
increasing number of threads. On all architecture, we report
performance per core for multi-threaded implementations normalized to
the performance attained on one thread.  Notice that the performance
per core for existing matrix-multiplication based convolutions
decrease significantly as we increase the number of threads. This is
an indication that as we increase the number of threads, the
processors are utilized less efficiently by the existing
matrix-multiplication based implementations.  Our direct CNN
implementation demonstrates minimal drop in performance per core as we
increase the number of threads. It is only when the number of threads
is twice as much as the number of physical cores does the performance
per core of our implementation drops significantly. This is expected
and important as it indicates that our implementation utilizes the
compute units effectively and increasing the number of threads beyond
the number of physical compute units creates excessive contention for
the compute resources, thereby resulting in a sharp drop in
performance per core.

\vspace{-2mm}
\section{Conclusions}
    \label{sec:conclusion}
In this paper, we demonstrate that direct convolution, a computational
technique largely ignored for computing convolution layers, is
competitive with existing state of the art convolution layer
computation.  We show that a high performance direct convolution
implementation not only eliminates all additional memory overhead, but
also attains higher performance than the expert-implemented
matrix-matrix-multiplication based convolution.  We also show that our
implementation scales to larger number of processors without
degradation in performance as our implementation exploits the
dimension of the kernel that has the highest amount of parallelism. In
contrast, current high performance matrix-multiply based
implementations do not scale as well to a larger number of processors.




Our direct convolution implementation currently attains 87.5\%, 58.2\%
and 88.9\% of the theoretical peak of the Intel, AMD, and ARM architecture, where as
the \texttt{SGEMM} on HPC matrices attains peaks of 89\% 54\% and 92\% the same architecture.
While we have shown that our direct convolution implementation is competitive (within 3\% of peak \texttt{SGEMM} performance), we believe that the performance gap between our direct convolution, and \texttt{SGEMM} on HPC matrices can be closed by taking an auto-tuning~\cite{PHiPAC97, ATLAS} or analytical
approach~\cite{Yotov:2005,BLIS4} to identifying the blocking
parameters of the different loops. 
These approaches will also allow the
exploration of different combinations of parallelism to determine suitable
parallelism strategies. This is something we intend to pursue in the
near future.


Another possible direction arising from this work is to use similar
design techniques to optimize the backward process to update both in
image and kernel. Given the similarity of the forward and backward
process, we believe that only minor changes to the loop ordering are
required.

Finally, we believe that our direct convolution algorithm can be
ported to the GPU. Our proposed data layouts are
similar to the layout required for the \texttt{StridedBatchedGemm} operation~\cite{shi2016tensor}. As 
this operation and data layout is currently supported on Nvidia GPUs using cuBLAS 8.0~\cite{cuBLAS}, this lends support 
to our belief that our algorithm can be easily ported to the GPU.

\section*{Acknowledgement}
This work was sponsored by NSF through award 1116802
and DARPA under agreement HR0011-13-2-0007. The content,
views and conclusions presented in this document do not
necessarily reflect the position or the policy of NSF or DARPA.


\bibliography{references,biblio}
\bibliographystyle{icml2018}

\end{document}